\documentclass{article}

\usepackage{microtype}
\usepackage{graphicx}
\usepackage{subfigure}
\usepackage{booktabs} % for professional tables

\usepackage{amsmath,amsfonts,bm}

\def\eqref#1{equation~\ref{#1}}
\def\1{\bm{1}}

\DeclareMathAlphabet{\mathsfit}{\encodingdefault}{\sfdefault}{m}{sl}
\SetMathAlphabet{\mathsfit}{bold}{\encodingdefault}{\sfdefault}{bx}{n}

\newcommand{\latentS}{\widetilde{\mathcal{S}}}
\newcommand{\latents}{\Tilde{s}}

\usepackage{lipsum}
\usepackage{xcolor}
\usepackage{hyperref}
\usepackage{url}
\usepackage{graphicx}
\usepackage[font=small]{caption}
\usepackage{wrapfig}
\definecolor{encoder}{RGB}{21,147,255}
\definecolor{dynamics}{RGB}{59,179,21}
\definecolor{predictor}{RGB}{255,21,230}
\definecolor{decoder}{RGB}{255,0,0}

\usepackage{hyperref}

 \usepackage[accepted]{icml2021}

\icmltitlerunning{Visualizing MuZero Models}
\begin{document}

\twocolumn[
%\icmltitle{Visualizing Learned MuZero Models}

%\icmltitle{Visualizing  Value Equivalence in Model-Based Reinforcement Learning}

\icmltitle{Visualizing MuZero Models}
%Visualizing Abstract MDPs in Model-Based Reinforcement Learning
%}




\begin{icmlauthorlist}
\icmlauthor{Joery A. de Vries}{lei,equal}
\icmlauthor{Ken S. Voskuil}{lei,equal}
\icmlauthor{Thomas M. Moerland}{lei}
\icmlauthor{Aske Plaat}{lei}
\end{icmlauthorlist}

\icmlaffiliation{lei}{Leiden Institute of Advanced Computer Science, Leiden, The Netherlands}
\icmlaffiliation{equal}{Authors contributed equally}

\icmlcorrespondingauthor{Joery A. de Vries}{j.a.de.vries.3@umail.leidenuniv.nl}
\icmlcorrespondingauthor{Ken S. Voskuil}{k.s.voskuil@umail.leidenuniv.nl}

\icmlkeywords{Model-based reinforcement learning, value equivalent models, representation learning, latent models}

\vskip 0.3in
]



\printAffiliationsAndNotice{}  % leave blank if no need to mention equal contribution

\newcommand{\fix}{\marginpar{FIX}}
\newcommand{\new}{\marginpar{NEW}}

\begin{abstract}

MuZero, a model-based reinforcement learning algorithm that uses a value equivalent dynamics model, achieved state-of-the-art performance in Chess, Shogi and the game of Go. In contrast to standard forward dynamics models that predict a full next state, value equivalent models are trained to predict a future value, thereby emphasizing value relevant information in the representations. While value equivalent models have shown strong empirical success, there is no research yet that visualizes and  investigates what types of representations these models actually learn. Therefore, in this paper we visualize the latent representation of MuZero agents. We find that action trajectories may diverge between observation embeddings and internal state transition dynamics, which could lead to instability during planning. Based on this insight, we propose two regularization techniques to stabilize MuZero's performance. Additionally, we provide an open-source implementation of MuZero along with an interactive visualizer of learned representations, which may aid further investigation of value equivalent algorithms.

\end{abstract}

\section{Introduction} \label{sec_introduction}
%\input{Chapters/intro}
% Muzero is new state of the art model based RL algorithm
% Plans on a learned model
% unclear what the learned model represents, not much research in understanding this

Model-based reinforcement learning has shown strong empirical success in sequential decision making tasks, as illustrated by the AlphaZero  \citep{silver_general_2018} and MuZero algorithms \citep{schrittwieser_mastering_2020}. Both of these approaches nest a planning loop, based on Monte Carlo Tree Search \citep{kocsis2006bandit,browne2012survey}, inside a learning loop, where we approximate global value and policy functions. While AlphaZero used a known model of the environment, MuZero  learned the model from sampled data. However, instead of a standard forward dynamics model, which learns to predict future states, the MuZero dynamics model is trained to predict {\it future values}, better known as a {\it value equivalent model} \citep{grimm_value_2020}.

A potential benefit of value equivalent models, compared to standard forward models, is that they will emphasize value and reward relevant characteristics in their representation and dynamics. This may be beneficial when the true dynamics are complicated, but the value relevant aspects of the dynamics are comparatively simple. As a second benefit, we train our model for its intended use: predicting value information during planning. Several papers have empirically investigated this principle in recent years \citep{tamar2016value,oh_value_2017,farquhar2018treeqn,silver_predictron_2017,schrittwieser_mastering_2020}, while \citet{grimm_value_2020} provides a theoretical underpinning of this approach. 

However, no literature has yet investigated what kind of representations these approaches actually learn, i.e., \emph{how} the learned representations are organized. The goal of this paper is therefore to investigate and visualize environment models learned by MuZero. Most interestingly, we find that, after training, an action trajectory that follows the forward dynamics model usually departs from the learned embedding of the environment observations. In other words, MuZero is not enforced to keep the state encoding and forward state prediction congruent. Therefore, the second goal of this paper is to regularize MuZero's dynamics model to improve its structure. We propose two regularization objectives to add to the MuZero objective, and experimentally show that these may indeed provide benefit. 

In short, after introducing related work (Sec. \ref{sec_related}) and necessary background on the MuZero algorithm (Sec. \ref{sec_background}), we discuss two research questions: 1) what type of representation do value equivalent models learn (Sec. \ref{sec:understanding}), and 2) can we use regularization to better structure the value equivalent latent space (Sec. \ref{sec:regularization})? We experimentally validate the second question in Sec. \ref{sec:setup} and \ref{sec:results}. Moreover, apart from answering these two question, we also open source modular MuZero code including an interactive visualizer of the latent space based on principal component analysis (PCA), available from \url{https://github.com/kaesve/muzero}. We found the visualizer to greatly enhance our understanding of the algorithm, and believe visualization will be essential for deeper understanding of this class of algorithms.

\section{Related Work} \label{sec_related}
Value equivalent models, a term introduced by \citet{grimm_value_2020}, are usually trained on end-to-end differentiable computation graphs, although the principle would be applicable to gradient-free optimization as well. Typically, the unrolled computation graph makes multiple passes through a dynamics model, and eventually predicts a value. Then, the dynamics model is trained through gradient descent on its ability to predict the correct value. The first value equivalent approach were Value Iteration Networks (VIN) \citep{tamar2016value}, where a differentiable form of value iteration was embedded to predict a value. Other variants of value equivalent approaches are Value Prediction Networks (VPN) \citep{oh_value_2017}, TreeQN and ATreeC \citep{farquhar2018treeqn}, the Predictron \citep{silver_predictron_2017}, and MuZero \citep{schrittwieser_mastering_2020}. These methods differ in the way they build their computation graph, where VINS and TreeQN embed entire policy improvement (planning) in the graph, VPNs, the Predictron and MuZero only perform policy evaluation. Therefore, the latter approaches combine explicit planning for policy improvement, which in the case of MuZero happens through MCTS.  \citet{grimm_value_2020} provides a theoretical analysis of value equivalent models, showing that two value equivalent models give the same Bellman back-up. 

MuZero uses the learned value equivalent model to explicitly plan through Monte Carlo Tree Search \citep{kocsis2006bandit,browne2012survey}, and uses the output of the search as training targets for a learned policy network. This idea of iterated planning and learning dates back to Dyna-2 \citep{silver2008sample}, while the particularly successful combination of MCTS and deep learning was introduced in AlphaGo Zero \citep{silver_mastering_2017} and Expert Iteration (ExIt) \citep{anthony2017thinking}.  In general, planning may add to pure (model-free) reinforcement learning: 1) improved action selection, and 2) improved (more stable) training targets. On the other hand, learning adds to planning the ability to generalize information, and store global solutions in memory. For more detailed overviews of value equivalent models and iterated planning and learning we refer to the model-based RL surveys by \citet{moerland2020model,plaat2020model}. 

Visualization is a common approach to better understand machine learning methods, and visualization of representations and loss landscapes of (deep) neural networks has a long history \citep{bischof1992visualization,yosinski2015understanding,karpathy2015visualizing,li2018visualizing}. For example, \citet{li_visualizing_2018} shows how the loss landscape of a neural network can indicate smoothness of the optimization criterion. Visualization is also important in other areas of machine learning, for example to illustrate how kernel-methods project low dimensional data to a high dimensional space for classification \citep{szymanski_visualising_2011}. Note that the most common approach to visualize neural network mappings is through \emph{non-linear} dimensionality reduction techniques, such as Stochastic Neighbour Embedding \cite{hinton_stochastic_2002}. We instead focus on linear projections in low dimensional environments, as non-linear dimensionality reduction has the risk of altering the semantics of the MDP models.

\section{The MuZero Algorithm} \label{sec_background}
We briefly introduce the MuZero algorithm \citep{schrittwieser_mastering_2020}. We assume a Markov Decision Process (MDP) specification given by the tuple $\langle \mathcal{S}, \mathcal{A}, \mathcal{T}, U, \gamma \rangle$, which respectively represent the set of states ($\mathcal{S}$), the set of actions ($\mathcal{A}$), the transition dynamics mapping state-action pairs to new states ($\mathcal{T}:\mathcal{S} \times \mathcal{A} \to p(\mathcal{S})$), the reward function mapping state-action pairs to rewards ($U:\mathcal{S} \times \mathcal{A} \to \mathbb{R})$, and a discount parameter ($\gamma \in [0,1]$) \citep{sutton_reinforcement_2018}. Internally, we define an abstract MDP $\langle \latentS, \mathcal{A}, \widetilde{\mathcal{T}}, R, \gamma \rangle$, where $\latentS$ denotes an abstract state space, with corresponding dynamics $\widetilde{\mathcal{T}}: \latentS \times \mathcal{A} \to \latentS$, and reward prediction $R: \latentS \times \mathcal{A} \to \mathbb{R}$. Our goal is to find a policy $\pi: \mathcal{S} \to p(\mathcal{A})$ that maximizes the value $V(s)$ from the start state, where $V(s)$ is defined as the expected infinite-horizon cumulative return:  

\begin{equation}
V(s) = \mathbb{E}_{\pi,T} \Big[ \sum_{t=0}^\infty \gamma^t \cdot u_{t} | s_0 = s  \Big].
\end{equation}

\begin{figure}[t]
\centering
    \includegraphics[width=.65\linewidth]{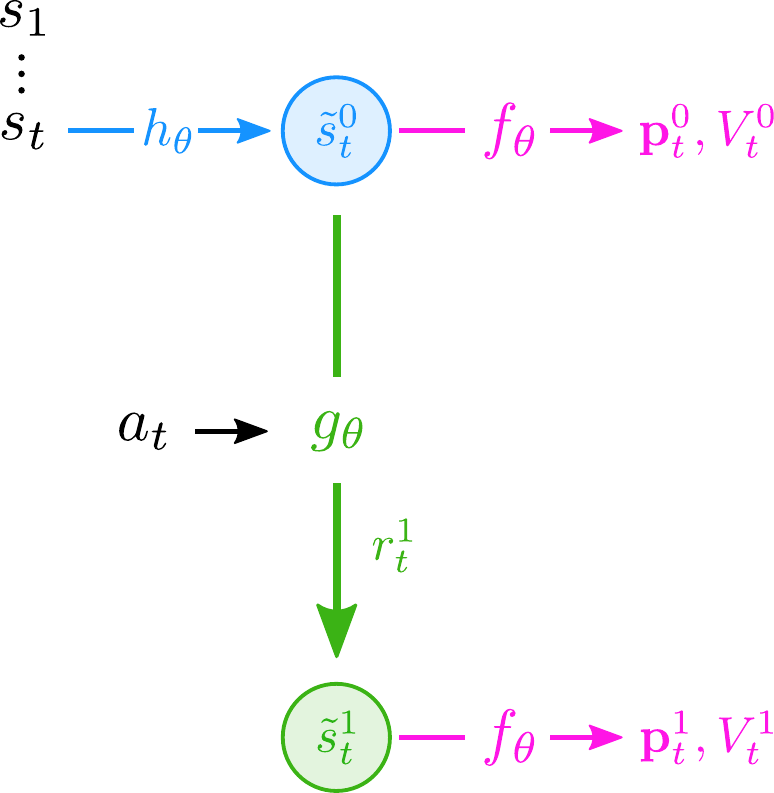}
    \caption{Illustration of MuZero's RNN unfolding during MCTS. During training, actions are given by the data.} \label{fig:background:MuMCTS} %\vspace{-3em}
\end{figure}

We define three distinct neural networks to approximate the above MDPs (Figure \ref{fig:background:MuMCTS}): the state encoding/embedding function $h_\theta$, the dynamics function $g_\theta$, and the prediction network $f_\theta$, where $\theta$ denotes the joint set of parameters of the networks. The encoding function $h_\theta: \mathcal{S} \to \latentS$ maps a (sequence of) real MDP observations to a latent MDP state. The dynamics function $g_\theta: \latentS \times \mathcal{A} \to \latentS \times \mathbb{R}$ predicts the next latent state and the associated reward of the transition. In practice, we slightly abuse notation and also write $g_\theta$ to only specify the next state prediction. Finally, the prediction network $f_\theta : \latentS \to p(\mathcal{A}) \times \mathbb{R}$ predicts both the policy and value for some abstract state $\latents$. We will identify the separate predictions of $f_\theta(\latents_t^k)$ by $\textbf{p}^k_t$ and $V^k_t$, respectively, where subscripts denote the time index in the true environment, and superscripts index the timestep in the latent environment. Also, we write $\mu_\theta = (h_\theta, g_\theta, f_\theta)$ for the joint model. 

Together, these three networks can be chained to form a larger computation graph that follows a single trace, starting from state $s_t$ following action sequence ($a_t,\dots,a_{t+n}$). First, we use the embedding network to obtain the first latent state from the sequence of observations: $\latents_t^0 = h_\theta(s_0, \dots, s_t)$. Then, we recursively transition forward in the latent space following $( \latents_t^{k+1}, r_t^{k+1}) = g_\theta(\latents_t^k, a_{t+k})$. Finally, for all generated abstract states ($\latents_t^0, \dots, \latents_t^n$) we use the prediction network to obtain $(\textbf{p}^k_t$, $V^k_t) = f_\theta(\latents_t^k)$. Altogether, the entire differentiable computation graph implements the mapping 

\begin{equation}
(s_0,..,s_t,a_t,..,a_{t+n}) \to (\textbf{p}^0_t, V^0_t, r_t^1,..,\textbf{p}^{n+1}_t, V^{n+1}_t, r_t^{n+1}), \label{eq_computation_graph}
\end{equation}

as illustrated in Figure \ref{fig:background:MuMCTS}. We will later use this graph to train our networks, but first discuss how they may be utilized to perform a search. 

Given three individual neural networks for $f$, $g$ and $h$ as, we then use these networks to perform a MCTS search from state $s_t$ in the abstract space. First, we embed the current state through $h$. Then, we can essentially follow the PUCT \citep{rosin_multiarmed_2011} algorithm, a variant of MCTS that incorporates prior weights on the available actions, where the priors originate from the policy network, $\textbf{p}^k_t$ (first proposed by \citet{silver_mastering_2017}). The state transitions and policy/value predictions within the search are governed through $g$ and $f$, respectively. Eventually, this MCTS procedure outputs a policy $\pi_t = \pi(s_t)$ and value estimate $V_t = V(s_t)$ for the root node, i.e., $(\pi_t,V_t) \sim \text{MCTS}(s_0, \dots, s_t|\mu_\theta)$. It then selects action $a_t \sim \pi_t$ in the real environment, transitions to the next state, and repeats the search. We refer to appendix B of the MuZero paper for full details on the MCTS search \citep{schrittwieser_mastering_2020}.

We still need to discuss how to train $f_\theta$, $g_\theta$, and $h_\theta$. Through playing episodes, MuZero collects trajectories $\eta$ of state, action, reward sequences, and the associated MCTS statistics obtained at each state, i.e., $\eta = (s_t,\pi_t,a_t,r_t,s_{t+1},\dots,\pi_{t+n},a_{t+n},r_{t+n},s_{t+n+1})$. As mentioned before, given a start state and a sequence of actions, we can unroll the differentiable graph that predicts rewards, values and policy parameters at each predicted state in the sequence (Eq. \ref{eq_computation_graph}, Fig \ref{fig:background:MuMCTS}). Since the entire graph is end-to-end differentiable, we may train it on a joint loss \citep{schrittwieser_mastering_2020}:

\begin{equation}\label{eq:background:loss}
\begin{split}
    l(\theta) = & \sum^n_{k=0} l^r(u_{t+k}, r^k_t) + l^v(z_{t+k}, V^k_t) + \\
    & l^p (\pi_{t+k}, \textbf{p}_t^k) + \lambda \|\theta\|^2_2,
    \end{split}
\end{equation}

where $l^r=0$ for $k=0$, $\lambda \in \mathbb{R}$ is a constant that governs the L2 regularization, and $z_t$ is a cumulative reward estimate obtained from an $n$-step target on the real trace $\eta$: 

\begin{equation} 
z_t = \sum^{n}_{i=0} \gamma^{i} u_{t+i} + \gamma^n V_{t+n+1}.
\end{equation}

For $n \to \infty$ this of course becomes a Monte Carlo return estimate. Moreover, MuZero uses distributional losses, inspired by \citet{pohlen_observe_2018, kapturowski_recurrent_2018}, for all three predictions ($r$, $V$, and $\textbf{p}$). 

Finally, MuZero employs two forms of normalization to stabilize learning. First, inside the MCTS selection rule, it min-max normalizes the value estimates based on the minimum and maximum estimate inside the current tree. This allows the algorithm to adapt to arbitrary reward scaling. Second, and more important for our paper, it also uses min-max normalization as the output activation of $h_\theta$ and $g_\theta$. In particular, every abstract state is readjusted through

\begin{equation}
\latents_\text{scaled} = \frac{\latents - \min(\latents)}{\max(\latents) -\min(\latents)},  
\end{equation}

where the $\min$ and $\max$ are over the element in the $\latents$ vector. This brings the range of $\latents$ in the same range as the discrete actions (one-hot encoded), but also introduces an interesting restriction, especially when the latent space is small. For example, when $| \latents|=4$, then the min-max normalization will enforce one of the elements of $\latents$ to be 0, and one of them to be 1. We will later see why this is relevant for our work. Full details on the MuZero algorithm can be found in the supplementary material of \citet{schrittwieser_mastering_2020}.

\section{Visualizing MuZero's Latent Space}  \label{sec:understanding}
\begin{figure*}[t]
    \centering
    \includegraphics[width=.9\linewidth]{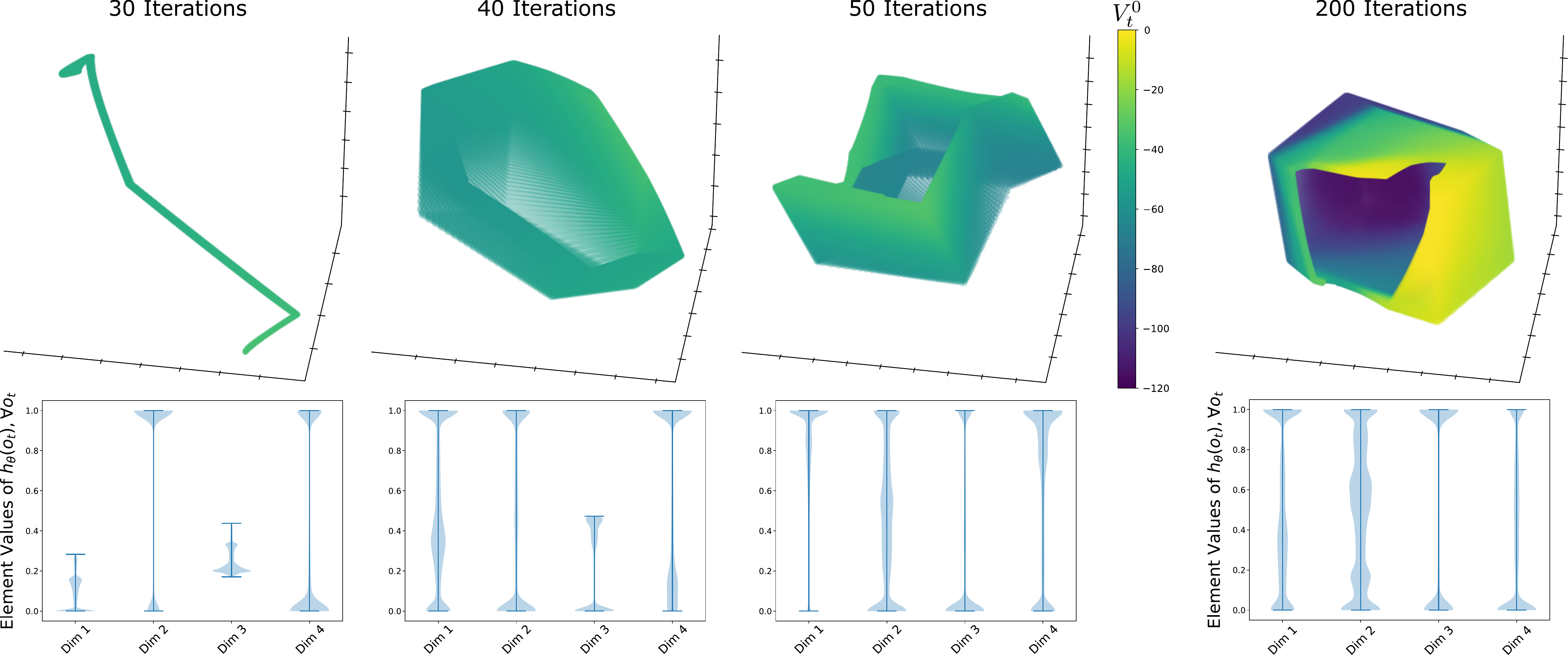}
    \caption{Development of the MuZero latent space over training. The progression on the top shows the 3D latent space, obtained by performing a principal component analysis on the agents true four-dimensional latent space. The color map indicates the learned value for each latent state. The bottom progression shows violin plots for the original 4D space, indicating which parts of the latent space are actually occupied during training.}
    \label{fig:results:embedding:mountaincar}
\end{figure*}

Value equivalent models, like MuZero, are solely trained on a low-dimensional, scalar reward signal, as the value and policy targets are also derived from the rewards. While the concept of  value equivalent models is convincing, it remains unclear what types of representations these models actually achieve, and how they shape their internal space. Therefore, we first train MuZero agents to subsequently map their embedding representation $\latents_t^0$ back to a 3D space, which we can graphically depict. 

\begin{figure*}[t]
    \centering
    \includegraphics[width=.8\linewidth]{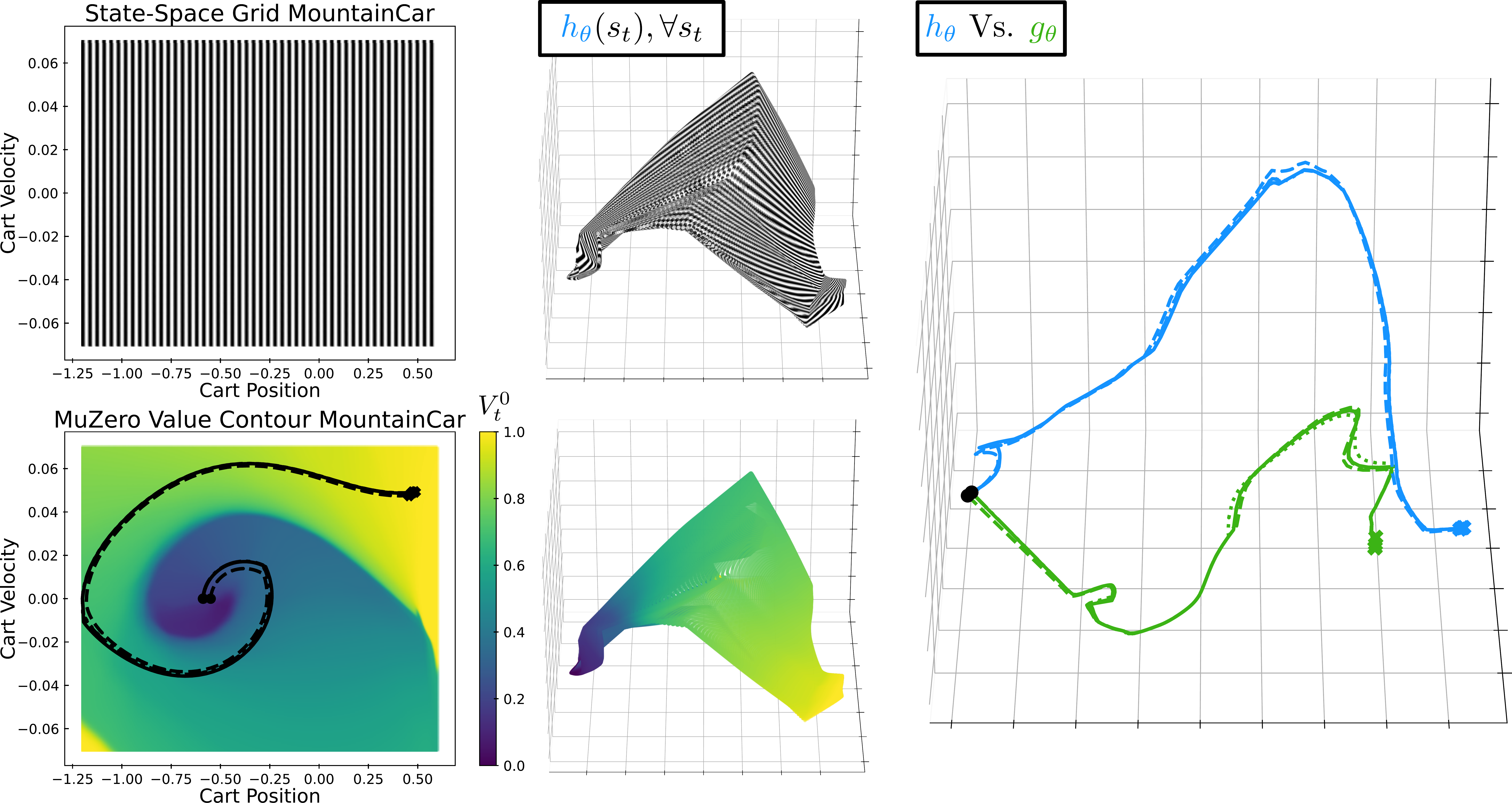}
    \caption{Left: Relation between true state space and latent state space on MountainCar. The top-left maps the true state space, indexed by vertical lines, into the latent space, retaining the vertical lines. The pattern indicates that the latent space gets curved, stretched and compressed. The bottom-left shows the same mapping, but this time indexes the states by their value estimate near convergence. We see that the latent embedding groups states that are dynamically close, which is visible in the grouping of states with similar values (in MountainCar, states that are dynamically close also have similar value, since there is only one start region and one sparse goal). Right: Latent embedding of a ground truth trace pushed through the encoder $h$ (blue) and latent embedding of the same trace embedded at the first state and subsequently propagated at latent level through $g$. We clearly see that, although both traces should cover the same trace and cumulative reward, their latent trajectories  diverge. }
    \label{fig:results:mdp:mountaincar}
\end{figure*}

Figure \ref{fig:results:embedding:mountaincar} illustrates a typical training progression for MuZero on the MountainCar environment (hyperparameters can be found in the appendix), where we use $L=|\latents|=4$. The top row shows the progression of the embedding $\{ \latents : \latents = h_\theta(s), \forall s \in \mathcal{S} \}$, shown in a 3D space obtained through PCA. Upon initialization, the latent space is almost degenerate, and all latent traces occupy the same part of the latent space. The action selection strategy of MuZero is near uniform in this situation. Gradually during training, the agent starts to expand and eventually occupies the full latent space. In the top-right plot, we also see that the latent spaces aggregate states with similar values, a topic to which we return later. On the bottom, we display violin plots of the latent state distribution during training, in the true four dimensional latent space. Most interesting, we see that these distributions appear overinflated around 0 and 1, on all dimensions. This is an artifact of the min-max normalization of the latent space (see end of previous section), which always forces one element to be 0, and one element to be 1. We do see that MuZero spreads its assignment of min and max over all dimensions. Note that, although MuZero can theoretically still predict any point in the latent space $[0,1]^4$, in practice the normalization  likely reduces the effective degrees of freedom within the latent space. 

Figure \ref{fig:results:mdp:mountaincar} further explores the latent space near algorithm convergence. The left part of the figure aims to relate the ground-truth state space to the latent space. On the top left, we visualize the ground-truth MountainCar state space (a 2D grid consisting of car position and car velocity) using vertical bars. Directly next to it, we visualize the associated MuZero latent space when moving through the embedding function, i.e., the subspace $\{ \latents : \latents = h_\theta(s), \forall s \in \mathcal{S} \}$. We retain the vertical bars, which thereby illustrates the way the true state space is curved into a high dimensional manifold. We do see that MuZero  recovers the state space to a large extent, while it was never explicitly trained on it, although some parts are compressed or stretched. 

The bottom-left of Figure \ref{fig:results:mdp:mountaincar} provides a similar illustration of the mapping between observed states and embedded states, but this time with accompanying value information. The bottom-left figure shows the learned value function of the state-space, where the black lines illustrate some successful final MuZero trajectories. Interestingly, when we map the value information into the embedding, we see that MuZero has actually strongly warped the original space. Relative to this projection, states with low value are grouped in the left-side of this space, while states with high value appear near the right-side. Effectively, MuZero has created a representation space in which it groups states that are dynamically close. For example, in the true state space (left), dark blue (low value) and yellow (high value) states are often adjacent. However, in the embedding (right), these states are strongly separated, because they are \emph{ dynamically distant} (we cannot directly move from blue to yellow, but need to follow the trajectory shown in black). This shows that MuZero indeed manages to retrieve essential parts of the dynamics, while only being trained on a scalar reward signal. 

Our last visualization of the learned model is shown in the right panel of Figure  \ref{fig:results:mdp:mountaincar}. In this plot, we again visualize a PCA of MuZero fitted on the embedding near convergence. However, this time we fix a start state ($s_0$) and action sequence ($a_0,\dots,a_n$), and visualize a successful trace passing through both the embedding and dynamics functions. For the embedding function, given a trace of environment data $\eta$, we plot $\langle h_\theta(s_0),h_\theta(s_1),\dots,h_\theta(s_{n+1}) \rangle$ as the \emph{blue} trajectory. In contrast, in \emph{green} we embed the first state and subsequently repeatedly transition through the learned dynamics function $g_\theta$---i.e., we plot $\langle h_\theta(s_0),g_\theta(h_\theta(s_0),a_0), g_\theta(g_\theta(h_\theta(s_0),a_0),a_1),\dots \rangle$. Interestingly, these two trajectories do not overlap. Instead, the latent forward dynamics seem to occupy a completely different region of the latent space than the state encoder. We repeatedly make this observation on different experiments and repetitions. Although one would conceptually expect these trajectories to overlap, there is no explicit loss term that enforces these trajectories to stay congruent. When the latent capacity is large enough, MuZero may fail to identify this overlap, which could waste generalization potential. Therefore, in the next section, we investigate two possible solutions to this problem.

\section{Regularizing the Latent Space} \label{sec:regularization}
\begin{figure}[t]
    \centering
    \includegraphics[width=\linewidth]{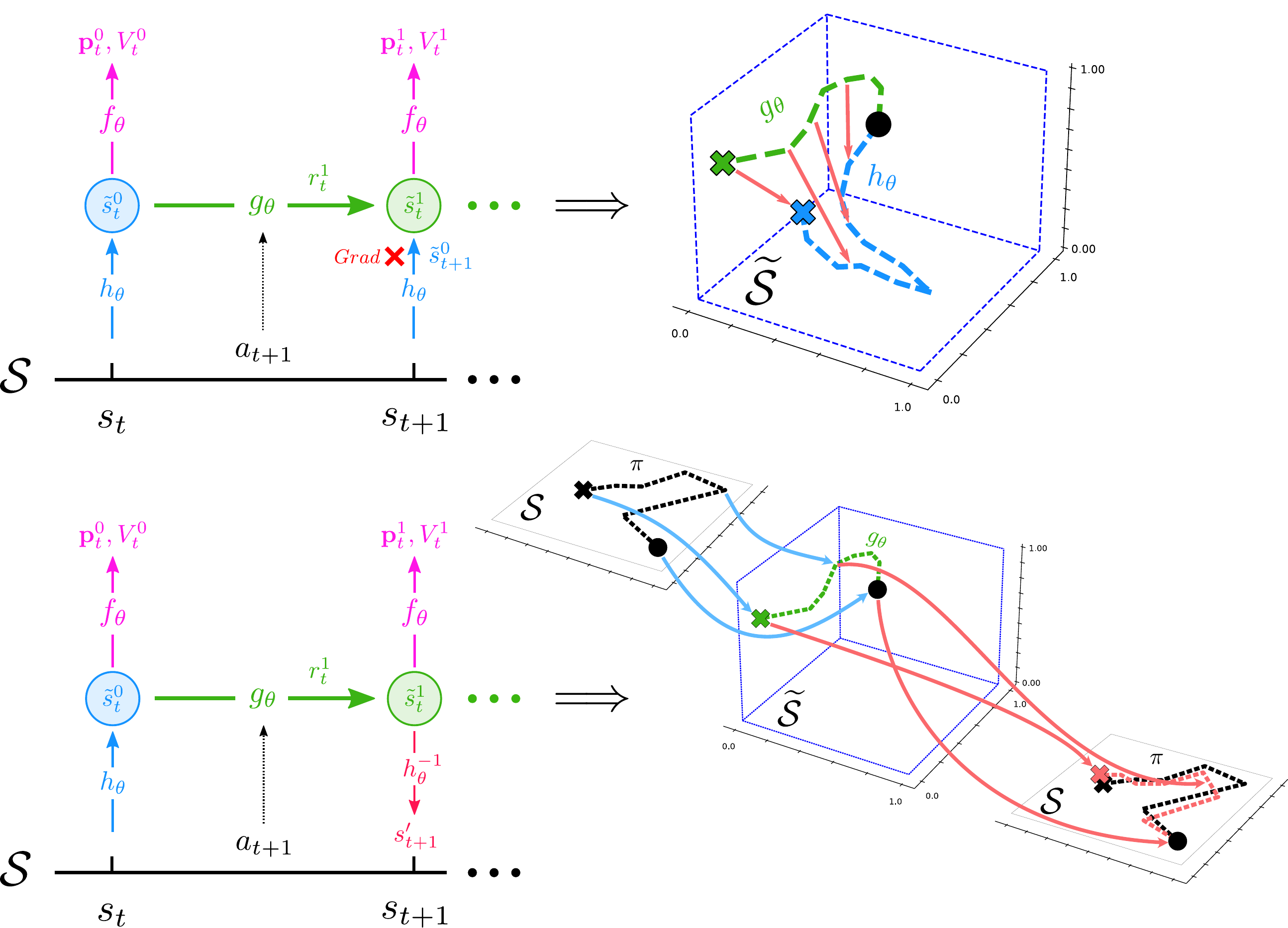}
    \caption{Two types of regularization of the MuZero objective. Top: {\it Contrastive regularization}, where we penalize the discrepancy between the embedding of a trace $h$ (blue dashed line) and the forward predictions of the same action sequence through $g$ (green dashed line). The contrastive loss pushes the green line towards the blue line (shown in red). Bottom: {\it Decoding regularization}, which adds an explicit decoding network $h^{-1}_\theta$, which maps predictions of the latent model $g$ (green dashed line) back to the observed state space (red lines). This enforces the learned latent model to retain some congruency with the true dynamics model.}
    \label{methods:fig:regularization}
\end{figure}

Ideally, the dynamics function should stay relatively congruent to the embedding of observed environment states. Thereby, we can maximally profit from generalization in the latent space, and make optimal use of the available data. However, as we observed in the previous section, there is no explicit training objective that ensures this congruence. We therefore propose two types of regularization, i.e., differentiable penalty terms (added to the loss) that enforce the latent space to be more congruent. 

The first solution is to add a {\it contrastive regularization}, which penalizes between the embedding of a state and the forward dynamics prediction towards the same abstract state. Given a trace $\eta$, we compute the loss

\begin{equation}
l^{c}(\theta) = \sum_{k=0}^n \| h_\theta(s_{t+k+1}) - \latents_t^{k+1}) \|_2^2, \label{eq_contrastive_reg}
\end{equation}

where $\latents_t^{k+1} = g(\dots g(g(h_\theta(s_0),a_0),a_1),\dots,a_k)$. This loss is added to the standard MuZero loss (Eq. \ref{eq:background:loss}), where its strength is governed by a scalar $\omega \in \mathbb{R}^+$. To prevent trivial solutions, we block the gradients of the above loss through $h_\theta$ for $k>0$. This ensures that only the dynamics function should move towards the embeddings (and not the other way around), assuming that the embedding function is already relatively well adapted. This type of contrastive loss is often seen in supervised learning, for example in Siamese Networks \cite{chen_exploring_2020, koch_siamese_2015}. The process is visually illustrated in the top graph of Figure~\ref{methods:fig:regularization}. 

A second solution can be the use of a {\it decoding regularization}, which decodes latent predictions back to the true observation. This essentially adds a (multi-step) forward prediction loss to the MuZero model. We initialize an additional decoder network $h^{-1}_\theta$, which maps latent states back to the true observations: $s'_{t+k} = h^{-1}_\theta(\latents_t^k)$. Then, given a trace of data $\eta$, we compute the loss 

\begin{equation}
l^{d}(\theta) = \sum_{k=0}^n \| s_{t+k+1} - h^{-1}_\theta(\latents_t^{k+1}) \|_2^2, \label{eq_decoder_reg}
\end{equation}

with $\latents_t^{k+1}$ defined as in the previous loss. Again, we can use a scalar $\omega \in \mathbb{R}^+$ to determine the relative contribution of the loss when added to Eq. \ref{eq:background:loss}. This approach combines the value equivalent loss model loss with a standard forward prediction loss. The additional penalty ties the latent predictions of $g_\theta$ back to the ground-truth observations. Although this mechanism is independent of the encoding $h$, it may potentially help to regularize the space of $g_\theta$ transitions. Moreover, as a second benefit, this regularization may be advantageous in initial stages of learning, as state transition are often dense in information compared to sparse reward signals. Thereby, it could be beneficial in early learning, but can be overtaken by the value equivalent objective when the model converges. The second type of regularization is visualized in the bottom graph of Figure \ref{methods:fig:regularization}.

\section{Experimental Setup} \label{sec:setup}
\begin{table}[!h]
    \centering
    \renewcommand{\arraystretch}{1.1}
    \caption{Overview of factorial experimental design: 1) regularization type, 2) regularization strength ($\omega$), 3) latent roll-out depth ($K$), and 4) latent dimensionality ($L = |\latents |$). Color annotations correspond to the loss regularization methods depicted in Figure \ref{methods:fig:regularization}.}
    \begin{tabular}{p{1.2cm}|p{2.9cm}|p{2.9cm}}
         Hyperpar. & CartPole & MountainCar \\
         \hline
         Regular. & $\{\mathrm{None},\textcolor{encoder}{\mathrm{Contr.}}, \textcolor{decoder}{\mathrm{Dec.}} \}$ & $\{\mathrm{None},\textcolor{encoder}{\mathrm{Contr.}}, \textcolor{decoder}{\mathrm{Dec.}} \}$ \\
         $\omega$ & $\{0.01, 0.1, 1\}$ & $\{1\}$ \\
         $K$ & $\{1, 5, 10\}$ & $\{3\}$ \\
         $L$ & $\{4, 8\}$ & $\{4, 8\}$ \\
    \end{tabular}
    \label{tab:methods:parameters}
\end{table}

We will now experimentally study our regularized versions of MuZero compared to standard MuZero. As an additional baseline, we also include an adaptation of AlphaZero for single-player domains \cite{silver_general_2018}, i.e., a similar agent with a perfect environment model. We provide a complete reimplementation of MuZero based on the original paper \cite{schrittwieser_mastering_2020}. All code is written in Python using Tensorflow 2.0 \cite{tensorflow2015-whitepaper}. The code follows the AlphaZero-General framework \cite{nair_suragnairalpha} and is available at \url{https://github.com/kaesve/muzero}, 
along with all resulting data. 

We tested our work on low-dimensional, classic control environments from  OpenAI's Gym \cite{openai_gym}, in particular CartPole and MountainCar. These are low-dimensional problems for which it is feasible to visualize and interpret the resulting model. We also developed an interactive visualization tool to visualize learned MuZero models on the MountainCar environment at \url{https://kaesve.nl/projects/muzero-model-inspector/#/}.\footnote{Please find a video of our tool in the supplementary material.}
We encourage the reader to play around with this tool as we personally found this to provide more insight than static $2$-D images.

All hyperparameters can be found in Appendix \ref{ap:parameters} in supplemental material. For the results in this paper, we report on a factorial experiment design shown in Table \ref{tab:methods:parameters}, where we vary: 1) the regularization type, 2) the regularization strength ($\omega$), 3) the number of latent unroll steps ($K$), and 4) the size of the latent representation ($L = |\latents|$). Of course, our main interest is in the ability of the regularizations to influence performance. All experiments were run on the CPU. The hyperparameter analysis experiment was run asynchronously on a computing cluster using 24 threads at 2GHz. %All MountainCar runs were run sequentially on a laptop with a 9th generation i7 CPU. 
On average it took about half a day to train an agent on any particular environment. The complete experiments depicted take about one week time to complete (after the separate lengthy tuning phase of the other hyperparameters).

\section{Results} \label{sec:results}
\begin{figure*}[t]
    \centering
    \includegraphics[width=.85\linewidth]{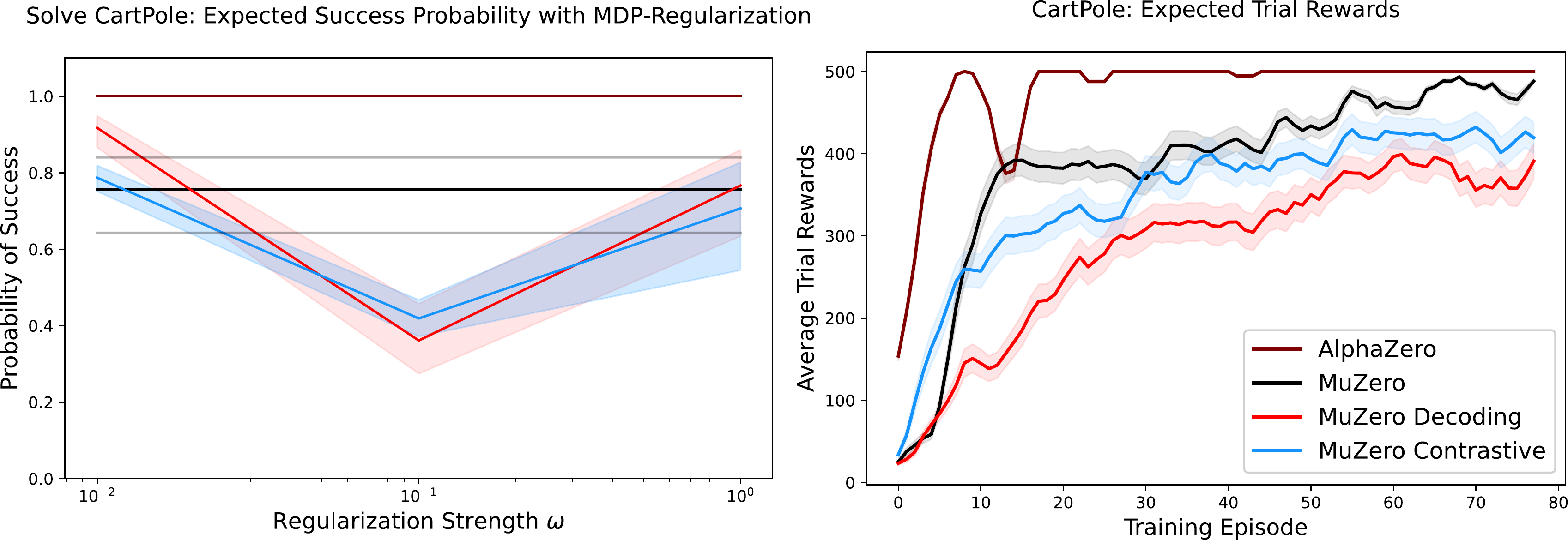}
    \caption{Performance of AlphaZero, MuZero and two regularized MuZero versions on CartPole. Results are averaged over all hyperparameters $K$ and $L$ (Table \ref{tab:methods:parameters}). Left: Success percentage as a function of the regularization strength. `Success' requires the maximum reward of $500$ consistently over $n=10$ test episodes. The error bars indicate a one standard error confidence interval of the mean success ratio.  Right: Learning curves for AlphaZero, MuZero and two regularized MuZero versions. These curves are averaged over all values of $K$ and $L$ in Table \ref{tab:methods:parameters}, which acts as a sensitivity analysis.}
    \label{fig:results:cartpole}
\end{figure*}

Figure \ref{fig:results:cartpole} shows results of our CartPole experiments. In both plots we compare AlphaZero, MuZero, and our two regularized MuZero extensions. On the left, we investigated the effect of the regularization strength. AlphaZero consistently solves this environment, while MuZero has more trouble and solves it roughly 75\% of the time. Both regularization methods do not seem to structurally improve MuZero performance on CartPole. Surprisingly, performance dips for intermediate regularization strength ($\omega = 0.1$). On the right of Figure \ref{fig:results:cartpole}, we display the average training curves for each of our four algorithms on CartPole. We see a similar pattern, where the regularization losses do not seem to provide benefit.

\begin{figure*}[t]
    \centering
    \includegraphics[width=.85\linewidth]{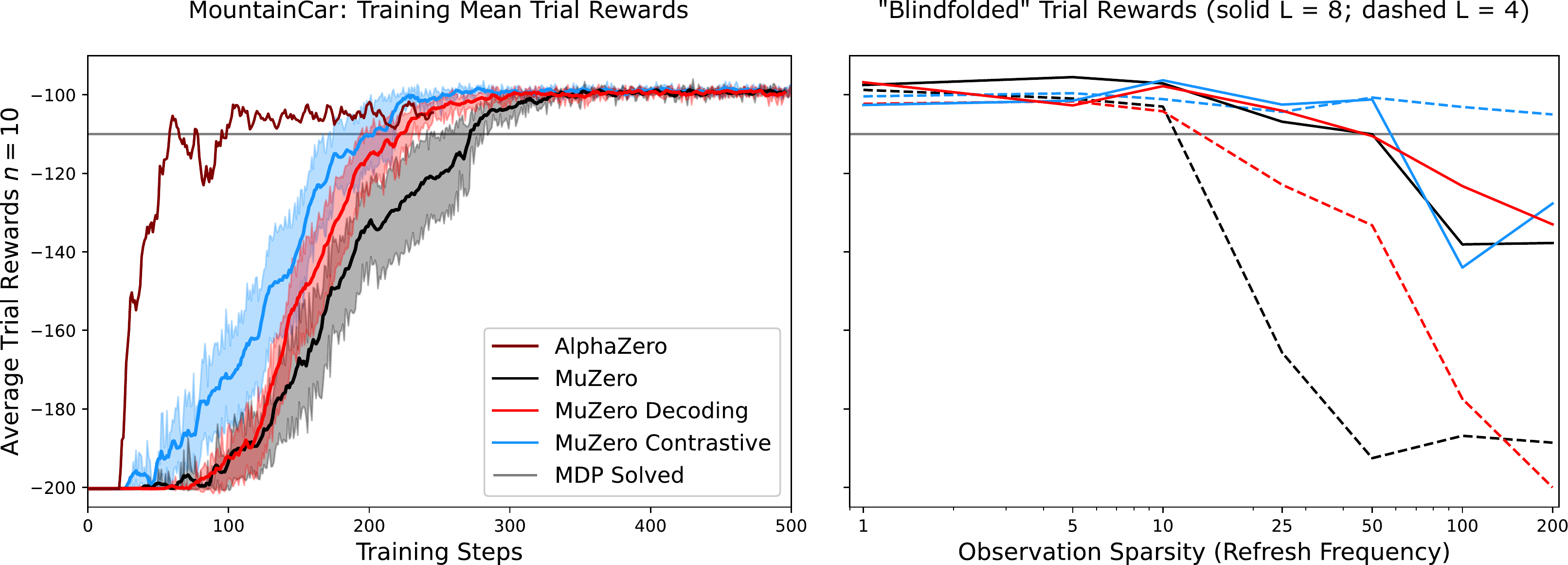}
    \caption{Performance of AlphaZero, MuZero and two regularized MuZero versions on MountainCar. Left: Reward progression during training. Results averaged over 10 repetitions. Right: Performance of blindfolded agents, where the agent only get to see the true environment observation every 'observation sparsity' number of steps. Solid curves for L=8, dashed curves for L=4. Results for a single repetition. }
    \label{fig:results:mountaincar}
\end{figure*}

Since CartPole is a dense reward environment, in which MuZero quickly obtains gradients to structure the latent space, we also investigate our reguralization methods on MountainCar. Figure \ref{fig:results:mountaincar} (left) shows the training progression for AlphaZero, MuZero and the two regularization methods. Again, AlphaZero is clearly the fastest to solve the environment. However, this time, both our regularization methods outperform MuZero in learning speed, where especially the contrastive regularization leads to faster learning. All methods do reach the same optimal performance. 

We further quantify the strength of the learned internal model by 'blindfolding' the agent \citep{freeman_learning_2019}. A blindfolded agent only gets to see the true state of the environment ever $x$ number of steps, where we call $x$ the `observation sparsity'. The right graph of Figure \ref{fig:results:mountaincar} displays the average reward obtained by all MuZero models when varying the observation sparsity (horizontal axis). All agents consistently keep solving the problem up to an observation sparsity of depth 10, which shows that their models do not compound errors too fast. Then, performance starts to deteriorate, although only the contrastively regularized, $L=4$ MuZero variant keeps solving the problem, even up to 200 steps without a ground truth observation. In general, the blindfolded agents show that the model stays consistent over long planning horizons (even when trained on shorter roll-out depths), and that especially contrastive regularization further increases this ability. Please refer to the Appendix for additional visualization of the latent space in MountainCar experiments.

\section{Discussion} \label{sec:discussion}
We studied the latent space of the MuZero algorithm, a successful combination of value equivalent models and iterated planning and learning. In particular, we visualized the latent space, showing how the true state space is partially recovered and warped into a dynamically relevant space. Moreover, we identified a frequent mismatch between the latent trajectories and the embeddings of the ground truth observations. To alleviate this issue, we introduced two forms of regularization, which did not improve performance in CartPole, but did improve performance on MountainCar. Finally, we also found that these models can accurately predict correct actions over long planning horizons, up to 200 steps in some cases, while being trained on relatively short trajectories.

We hypothesize that the benefit of regularization is most prominent on sparse reward tasks like MountainCar. In CartPole, where we quickly observe rewards, the representation space of MuZero will quickly receive relevant gradients to structure itself. However, when reward information is sparse, we may benefit from additional regularization of the latent space, which might squeeze additional information out of the available data. Especially the contrastive regularization seems to show this effect, and could be a promising direction for future research. 

A critique on our work could involve the dimensionality of the experiments. Deep reinforcement learning has started to focus more and more on high-dimensional problems. Nevertheless, we opted for small scale experiments for two reasons. First of all, low-dimensional problems allow for better interpretation. We can only visualize a latent space in three dimensions, and it would not make sense for example to map an Atari game representation back to 3D. Moreover, the general emphasis in RL has shifted towards to the challenge of dimensionality, while problems can be challenging in more ways than dimensionality only. A good illustration of the latter point is that such a high-impact algorithm as MuZero actually suffers to consistently solve CartPole and MountainCar. This recognition of the value of low dimensional experiments also starts to resurface in the RL community \citep{osband2019behaviour}. 

As a second reason, we  chose small-scale experiments because AlphaZero and MuZero are very computationally expensive to run. While standard RL algorithms are already expensive to run, these iterated search and learning paradigms are significantly more expensive. When we use a small-scale MCTS of $n=11$ at every timestep,  it already takes at least 11 times as long to complete the same number of episodes as model-free RL. 
%Note that, within  MCTS, we have to make many passes through the neural networks, which further increases computational complexity. 
Compared to model-free RL, these iterated search and learning paradigms seem to benefit in long run performance, but they are computationally heavy to train. %The experiments in this paper already take more than a week of computation, without all the hyperparameter scaling involved before. 

We  observed that AlphaZero and, especially, MuZero are rather unstable to train. Their performance strongly depends on a variety of hyperparameters and network initialization, and we observe strong variation in the performance over different runs. A direction for future work is to further stabilize these algorithms. Our work on novel regularizations for value equivalent models like MuZero is a first step in this direction. Future work could investigate how to further stabilize MuZero through different regularizations, or further combinations with standard forward prediction or types of auxiliary losses \citep{jaderberg2016reinforcement}. Although value equivalency is a strong incentive for the final representation of the model, we may aid the model during learning with additional losses and regularizations, especially in sparse reward tasks. 

% Add later?: Areas of high reward and low reward seem maximally separated, this may give an indication of why planning works. MuZero's dynamics seem to be able to plan over reward preferences (it does not have to rely on a root-level prediction as it can step forward throughout its model to evaluate multiple preferenes and arrive at a final decision.

% Add discussion: Density of data for observations and points within the learned space to visualize how the agent explores within both spaces.

% Add discussion: Dynamics upon refresh jumps back to original trajectory.

\section{Conclusion}
The first part of this paper visualized the learned latent representations and model of MuZero, which we consider the main contribution of the paper. Our two most important observations are: 1) the learned latent space warps the original space into a 'dynamically close' space, in which states that are easy to reach from eachother are nearby, and 2) the embedding of an action trajectory through the latent dynamics often departs from the embedding of the ground-truth states, which suggest the latent space does not optimally generalize information. To alleviate the last problem, the second part of the paper investigated two new forms of regularization of value equivalent models: 1) a contrastive regularization, and 2) a decoder regularization. Experiments indicate that these approaches may aid MuZero in sparse reward tasks, and that these models in general manage to plan over long horizons (much longer than they were trained on).
Our visualizations of the latent space indicate that e.g., the contrastive regularization term can force the dynamics function to predict values close to the embedding, this in turn can help interpreting the look-ahead search of MuZero. All code and data to reproduce experiments is open-sourced at \url{https://github.com/kaesve/muzero}, including an interactive visualizer to further explore the MuZero latent space. We also hope our work further stimulates new emphasis on visualization of RL experiments, which may provide more intuitive insights into deep reinforcement learning.

%\section*{Software}
%We provide an open-source modular framework for designing and applying AlphaZero and MuZero agents at
%\url{https://anonymized.org}.
%%\url{https://github.com/kaesve/muzero}.
%The AlphaZero algorithm has been adapted to work on both singleplayer domains and boardgames. The repository contains further details to reproduce our work along with a short developer's guide. We also provide an interactive visualization tool for investigating MuZero agents for the MountainCar environment at \url{https://anonymized.org}
%%\url{https://kaesve.nl/projects/muzero-model-inspector/}.

\bibliography{references}
\bibliographystyle{icml2021}

%\printbibliography

\newpage

\appendix
\section{Hyperparameters and Neural Architectures} \label{ap:parameters}
Our neural network architectures for MuZero were partially inspired by the ones proposed by \cite{van_seijen_loca_2020}. We kept every constituent network within the larger MuZero model equivalent, that is: $h_\theta$, $g_\theta$, $f_\theta$, and optionally $h^{-1}_\theta$ had the same \emph{internal} architecture being a two-layered feed forward network with $32$ hidden nodes using the Exponential Linear Unit activation function. The embedding function $h_\theta$ received only current observations, so no trajectories. Actions for the dynamics functions were one-hot-encoded and concatenated to the current latent state before being passed to $g_\theta$; computed latent-states were also minmax-normalized within the neural architecture. The action policy, value function, and reward prediction were represented as distributions and trained using a cross-entropy loss, as was also done in the original paper \cite{schrittwieser_mastering_2020}. The AlphaZero neural network was simply $f_\theta$ that received state observations---i.e., when omitting $h_\theta$ and $g_\theta$. All other unmentioned hyperparameters are shown in Table \ref{tab:appendix:parameters}.

\section{Additional Visualization of Latent Space}
\begin{figure*}[t]
    \centering
    \includegraphics[width=\linewidth]{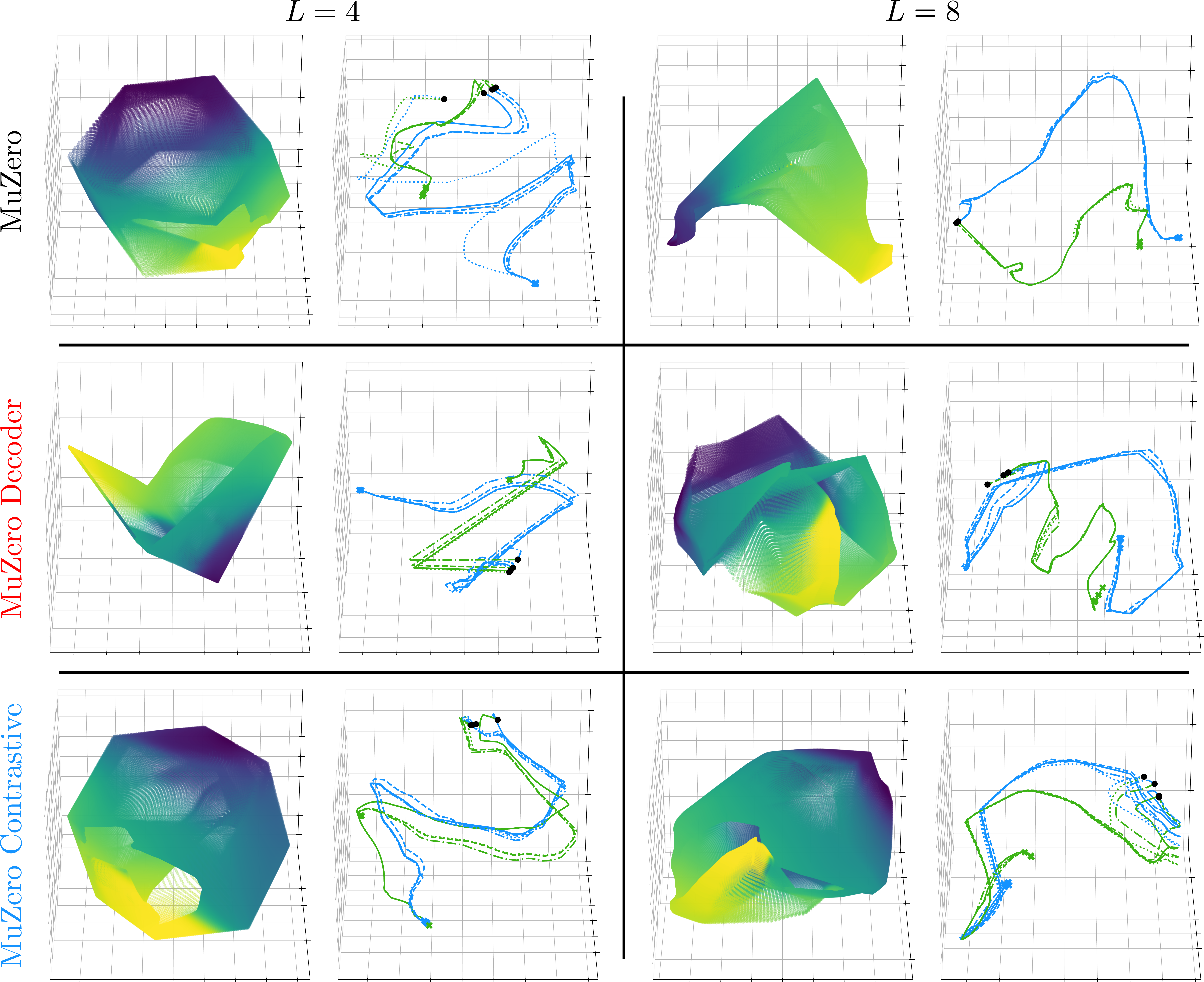}
    \caption{Rows: Different algorithms (MuZero, MuZero Decoder and MuZero Contrastive). Columns: Different latent sizes ($L=4$, $L=8$). For each combination we show two graphs: 1) (left) a 3-dimensional PCA projection of the learned embeddings in MountainCar and 2) (right) uniquely sampled trajectories embedded through $h_\theta$ (blue) or simulated forward through the latent dynamics $g_\theta$ (green). We see that both trajectories depart for standard MuZero, but especially contrastive regularization manages to keep these trajectories congruent.}% Interestingly, the $L=4$ agents' embeddings slightly resemble a polytope in the Principal Component projection.}
    \label{ap:figure:l8}
\end{figure*}

Figure \ref{ap:figure:l8} provides additional illustration of the learned latent spaces for MuZero and our two regularized MuZero variants on MountainCar. The figure has a 3x2 design, where the rows depict different algorithms, and the columns different sizes of the latent space. The left of each cell contains a 3-dimensional PCA projection of the learned latent space. We again observe that the value equivalent objective nicely unfolds a latent space, in which states that are dynamically close in the true environment are also close in the latent space. 

More interestingly, the right of each cell displays the two ways to embed a trajectory: 1) by passing each observed true state through $h_\theta$ (blue), or 2) by simulating forward through the latent dynamics $g_\theta$ (green). We again observe that these trajectories strongly depart for the original MuZero algorithm. In contrast, the two regularization methods force both trajectories to stay closer to eachother. Especially the contrastive loss manages to keep both trajectories close to consistent, which may help to obtain better generalization, and could explain the faster learning observed in Figure \ref{fig:results:mountaincar}.

\begin{table*}[t]
    \centering
    \renewcommand{\arraystretch}{1.3}
    \caption{Overview of all constant hyperparameters used in our experiments for both AlphaZero and MuZero, see Table \ref{tab:methods:parameters} for the non-constant hyperparameters.}
    \label{tab:appendix:parameters}
    \begin{tabular}{p{3cm}|l|p{8cm}}
         Hyperparameter & Value (CP/ MC) & Description\\
         \hline
         Self-Play Iterations & $80$/ $1000$ & Number of data-gathering/ backpropagation loops to perform. \\
         Episodes & 20 & Number of trajectories to generate within the environment for each Self-Play loop. \\
         Epochs & 40 & Number of backprop steps to perform for each Self-Play loop. \\
         Episode Length & 500/ 200 & Maximum number of steps per episode before termination. \\
         $\alpha$ & 0.25 & Dirichlet alpha used for adding noise to the MCTS root prior. \\
         $p_{explore}$ & 0.25 & Exploration fraction for the MCTS root prior. \\
         $c_1$ & 1.25 & Exploration factor in MCTS's UCB formula. \\
         $c_2$ & 19652 & Exploration factor in MCTS's UCB formula. \\
         $\tau$ & 1 & MCTS visitation count temperature. \\
         $N$ & 11 & Number of MCTS simulations to perform at each search. \\
         
         Prioritization & False & Whether to use prioritized sampling for extracting training targets from the replay buffer. \\
         Replay-Buffer window & 10 & Number of self-play loops from which data is retained, self-play iterations outside this window are discarded \\
         TD-steps & 10/ 50 & Number of steps until bootstrapping when computing discounted returns. \\
         $\gamma$ & 0.997 & Discount factor for computing $n$-step/ TD-step returns. \\
         
         Optimizer & Adam & Gradient optimizer for updating neural network weights. \\
         $\eta$ & $2\cdot 10^{-2}$ & Adam Learning Rate \\
         Batch Size & 128 & Number of target samples to use for performing one backpropagation step. \\
         $\lambda$ & $10^{-4}$ & L2-Regularization term for penalizing magnitude of the weights. \\
         $S$ & $15$/ $20$ & Number of distribution support points for representing the value and reward functions.
         
    \end{tabular}
\end{table*}

\end{document}